\documentclass[preprint,12pt,number]{elsarticle}

\biboptions{sort&compress}
\usepackage[a4paper,
            bindingoffset=0.2in,
            left=0.5in,
            right=0.5in,
            top=0.7in,
            bottom=0.7in,
            footskip=.25in]{geometry}

\usepackage{array}
\usepackage{booktabs}
\usepackage{multirow}
\usepackage{siunitx}
\usepackage{verbatim}
\usepackage{enumitem}
\usepackage[hidelinks]{hyperref}
\usepackage{adjustbox}
\usepackage{color,soul}
\usepackage{caption}
\usepackage{subcaption}
\usepackage{array}
\usepackage[dvipsnames]{xcolor}
\usepackage[table]{xcolor}

\usepackage[utf8]{inputenc}
\usepackage{mathtools, nccmath}

\usepackage{amsmath, amsfonts}
\usepackage{amssymb}
\usepackage[mathlines]{lineno}
\usepackage{lineno}

\usepackage{textcomp}
\usepackage{graphicx}
\usepackage{algorithm} 
\usepackage{algpseudocode}
\usepackage{algorithmicx}
\usepackage[figurename=Fig.]{caption}

\usepackage{wrapfig}
\DeclareCaptionLabelFormat{custom}{\textbf{#1 #2}}
\DeclareCaptionLabelSeparator{custom}{. }
\DeclareCaptionFormat{custom}{#1#2 #3}
\begin{document}

\begin{frontmatter}

\title{Fully Convolutional Spatiotemporal Learning for Microstructure Evolution Prediction}

\author[inst1]{{Michael Trimboli}}

\author[inst1]{{Mohammed Alsubaie}}

\author[inst2]{Sirani M. Perera}
\author[inst3]{Ke-Gang Wang\corref{cor1}}
\author[inst1]{Xianqi Li\corref{cor1}}

\cortext[cor1]{Corresponding authors}

\affiliation[inst1]{organization={Department of Mathematics and Systems Engineering},
            addressline={Florida Institute of Technology}, 
            city={Melbourne},
            state={FL},
            postcode={32901}, 
            country={USA}}

\affiliation[inst2]{organization={Department of Mathematics},
            addressline={Embry-Riddle Aeronautical University}, 
            city={Daytona Beach}, 
            state={FL},
            postcode={32114},
            country={USA}}

\affiliation[inst3]{organization={Department of Mechanical and Civil Engineering},
            addressline={Florida Institute of Technology}, 
            city={Melbourne},
            state={FL},
            postcode={32901}, 
            country={USA}}

\begin{abstract}
Understanding and predicting microstructure evolution is fundamental to materials science, as it governs the resulting properties and performance of materials. Traditional simulation methods, such as phase-field models, offer high-fidelity results but are computationally expensive due to the need to solve complex partial differential equations at fine spatiotemporal resolutions. To address this challenge, we propose a deep learning-based framework that accelerates microstructure evolution predictions while maintaining high accuracy. Our approach utilizes a \textit{fully} convolutional spatiotemporal model trained in a self-supervised manner using sequential images generated from simulations of microstructural processes, including grain growth and spinodal decomposition. The trained neural network effectively learns the underlying physical dynamics and can accurately capture both short-term local behaviors and long-term statistical properties of evolving microstructures, while also demonstrating generalization to unseen spatiotemporal domains and variations in configuration and material parameters. Compared to recurrent neural architectures, our model achieves state-of-the-art predictive performance with significantly reduced computational cost in both training and inference. This work establishes a robust baseline for spatiotemporal learning in materials science and offers a scalable, data-driven alternative for fast and reliable microstructure simulations.

\end{abstract}

\begin{keyword}
Microstructure Evolution; Deep Learning; Fully Convolutional Neural Network; Spatiotemporal Models; Phase-field Simulations
\end{keyword}

\end{frontmatter}


\section{Introduction}
The evolution of microstructures is a fundamental phenomenon that governs the formation and transformation of internal structures in materials over time and plays a fundamental role in determining the behavior of materials \cite{KGWang05}. The characteristics of microstructures, such as grain size \cite{LQChen02}, phase distribution \cite{Trieger23}, and interfacial morphology \cite{GSRohrer05, Zinke-Allmang99}, directly influence mechanical, thermal, electrical, and chemical properties \cite{MECates18, JCLi19}. As such, understanding and predicting microstructure evolution is essential to advancing the design and optimization of materials across a wide range of engineering applications, including metallic materials, ceramics, polymers, and biomaterials \cite{wangb1}.

Many engineering materials are inherently multiphase to fulfill specific functional requirements. For instance, technologically important materials such as superalloys and precipitation-hardened metals are composed of multiple constituent phases with distinct sizes, morphologies, and physical properties \cite{ADeschamps21}. The overall performance of these materials is strongly influenced by the spatial arrangement and characteristics of individual grains and phases. Moreover, interfaces such as grain boundaries and phase boundaries can exhibit vastly different mechanical and transport properties compared to the bulk material, further affecting behavior at the macroscopic level \cite{KGWang05}.

Microstructure evolution occurs through several key physical processes, including Ostwald ripening (phase coarsening) \cite{wangb1,KGWang052,KGWang06,KGWang10,HuiY,KGWang21,KGWang24,KGWang17}, grain growth \cite{Fradkov94}, spinodal decomposition \cite{YOono88}, and solidification \cite{WKurz21}. Each of these mechanisms is governed by coupled thermodynamic and kinetic principles and results in time-dependent changes in microstructural patterns. For example, grain growth leads to coarsening of the grain structure through boundary migration, while spinodal decomposition describes spontaneous phase separation driven by composition fluctuations. These complex transformations have been extensively studied through experimental, theoretical, and computational methods.

The phase-field methods have emerged as one of the most powerful computational frameworks for simulating microstructural dynamics. They have been broadly applied to investigate various important evolving microstructures that occur in grain growth and coarsening \cite{krill02acta}, solidification \cite{YZhao19}, thin-film deposition \cite{Stewart20}, dislocation dynamics \cite{JBeyerlein16}, and crack propagation \cite{ISAranson00}. By representing microstructures as continuous fields governed by nonlinear partial differential equations (PDEs), phase-field models are able to capture complex interfacial phenomena, topological transitions, and morphological evolution with high fidelity. As a result, they provide deep physical insight into processes such as grain growth, spinodal decomposition, and dendritic solidification. However, this accuracy typically comes at a \textit{high computational cost}, as phase-field simulations often require solving coupled PDEs over fine spatial and temporal resolutions for long physical times.

Efforts to mitigate this computational burden have traditionally focused on high-performance computing (HPC) architectures \cite{shimokawabe2011peta,hunter2011large,vondrous2014parallel,HuiY,EMiy,XShi} and advanced numerical techniques \cite{DJSeo,TMur,QDu}. Despite these approaches, achieving a practical balance between accuracy and computational efficiency remains a significant challenge, particularly for applications requiring repeated simulations, parameter sweeps, or long-time evolution. These limitations have motivated the exploration of data-driven surrogate models as complementary tools for accelerating microstructure evolution prediction. 
In recent years, deep learning has emerged as a promising alternative for approximating complex dynamical systems directly from data generated by physics-based solvers \cite{zhang2020machine,DOZapiain,oommen2022learning,hu2022accelerating,yang2021self,farizhandi2023spatiotemporal,ahmad2023accelerating}. A variety of neural network architectures have been investigated for microstructure evolution modeling, including recurrent neural networks (e.g., PredRNN++ \cite{wang2018predrnn++}), three-dimensional (3D) recurrent models such as E3D-LSTM \cite{wang2018eidetic}, operator-learning approaches including Fourier Neural Operators (FNO) \cite{li2020fourier}, physics-informed neural networks (PINNs) \cite{raissi2019physics}, and graph neural networks (GNNs) \cite{scarselli2008graph}. While each class of methods offers distinct advantages, they also present limitations when applied to long-horizon, high-resolution microstructure evolution problems. 

Recurrent-based architectures explicitly propagate hidden states through time and often demonstrate strong predictive capability. However, their sequential nature limits temporal parallelization and introduces computational overhead, which may lead to training instability and error accumulation over long prediction horizons \cite{hu2022accelerating,farizhandi2023spatiotemporal}. These issues are particularly critical in microstructure evolution, where small inaccuracies at grain boundaries or phase interfaces can amplify and produce qualitatively incorrect morphologies. Operator-learning approaches such as FNO aim to approximate solution operators in spectral space and have shown promise for certain classes of PDEs \cite{oommen2022learning}. Nevertheless, their reliance on global Fourier representations may be less effective for highly heterogeneous microstructures dominated by localized interfacial dynamics. Similarly, PINN-based methods incorporate governing equations directly into the loss function, offering a principled framework for enforcing physical constraints. However, PINNs are typically implemented using coordinate-based multilayer perceptrons and often encounter scalability challenges when applied to high-dimensional spatiotemporal fields with nonlinear phase-field dynamics \cite{kalesh2025application}. GNNs provide an alternative representation by modeling grains or microstructural features as nodes and their interactions as edges \cite{qin2024graingnn}. While this abstraction is well suited for capturing topological relationships at the grain level, it typically requires explicit graph construction and may discard fine-scale interfacial information present in pixel-resolved phase-field representations. Furthermore, dynamically evolving microstructures necessitate frequent graph updates, which can increase computational complexity and limit scalability for long-horizon prediction. 
In contrast, fully convolutional spatiotemporal models offer a streamlined and efficient alternative that operates directly on dense microstructure fields. By eliminating recurrent units, transformer blocks, and global spectral operators, the fully convolutional architecture enables parallel computation across both spatial and temporal dimensions, improving scalability and training stability. Localized convolutional and attention-based mechanisms naturally align with the physics of microstructure evolution, where dynamics are primarily driven by local interfacial interactions rather than global dependencies.

In this work, we investigate a fully convolutional spatiotemporal learning framework based on the SimVPv2 architecture \cite{tan2022simvpv2}. The proposed model is lightweight, non-recurrent, and free of Unet-like \cite{pham2024seunet} multi-scale hierarchies, allowing efficient multi-step prediction in a single forward pass. Trained in a self-supervised manner using phase-field simulation sequences of grain growth and spinodal decomposition, the network learns to approximate the underlying evolution operator directly from data. Numerical experiments demonstrate that the proposed approach accurately captures both short-term local transitions and long-term statistical characteristics of evolving microstructures, while maintaining high computational efficiency. Compared to recurrent-based spatiotemporal models, the fully convolutional architecture exhibits improved scalability to longer prediction horizons and diverse configurations. These results highlight the potential of fully convolutional spatiotemporal learning as an efficient and scalable complement to PDE-based simulation, establishing a strong baseline for data-driven microstructure evolution modeling and providing a pathway toward accelerated simulation and microstructure-informed materials design.

The remainder of this paper is organized as follows. In Section 2, we briefly describe the phase-field simulation framework. Section 3 presents the formulation and detailed description of the proposed deep spatiotemporal learning model for microstructure evolution prediction. The experimental setup is introduced in Section 4. Section 5 reports the results of the current study in detail. Finally, Section 6 provides a discussion and concluding remarks, summarizing the main findings of this work.

\section {Phase-field Method for the Study of Microstructure Evolution }
\label{sec:PFM}
In materials science, microstructure evolution occurs through diverse mechanisms and critically influences the properties of materials. Accurately and efficiently modeling it is thus essential to discover processing-microstructure-property relationships in materials design. Phase-field methods, developed since the early 1990s, have emerged as one of the most powerful computational tools for simulating microstructure evolution \cite{LQChen02}.  These methods use a diffuse-interface description, where spatial and temporal microstructure evolution is modeled through continuum equations such as the Cahn-Hilliard (CH) nonlinear diffusion equation \cite{JWCahn} and the Allen-Cahn (AC) (or time-dependent Ginzburg-Landau) equation \cite{JWCahn2}.
Over the last 30 years, phase-field simulations have been one of the most popular computational method used to study microstructure evolution in materials science,  computational mathematics, and physics. It has been extensively used to describe a variety of important evolutionary phenomena including grain growth \cite{krill02acta,chenwang01,kawang01,kwc01,schmidt}, spinodal composition \cite{wise}, phase coarsening \cite{wangb1, KGWang10, KGWang21,KGWang24,allen79}, solidification \cite{boettinger01,mathiesen2002time}, and vesicles formation in biological membranes \cite{LSta}. 

In general, for nearly every phase-field model, the key component is to formulate the free energy $F$ as a functional
\begin{equation}
F  =\int f(c_1, \cdots, c_n, \eta_1, \cdots, \eta_n, \nabla c_1, \cdots, \nabla c_n, \nabla \eta_1, \cdots, \nabla \eta_n, p, T, \cdots) dv,
\label{FreeE}
\end{equation}
where $c_i$ and $\eta_i$, $i=1, \cdots, n$ denote respective conserved and nonconserved field variables that are continuous across the interfacial regions \cite{Dtourret}. The energy density $f$ is typically a function of 1) a potential with local minima in the two (or more) phases, 2) gradient terms of $\nabla c_i$ and $\nabla \eta_i$ that are related with the energetic cost of interfaces, and 3) bulk energy density terms as a function of the local fields $c_i$ and $\eta_i$, local state variables such as pressure $p$ and temperature $T$, and other possible external factors. 

The temporal and spatial evolution of the field variables $c_i$ and $\eta_i$ are governed by the CH nonlinear diffusion equation and the AC relaxation equations, respectively,
\begin{equation}  
\frac{\partial c_i}{\partial
t}={\bf\nabla}\left\{{D_{c_i}\bf\nabla}\left [\frac{\delta F} {\delta c_i}\right ]\right\}; \qquad i=1,\dots n,
\label{CH}
\end{equation}

\begin{equation}
\frac{\partial \eta_i}{\partial
t}=-L_{\eta_i}\frac{\delta F} {\delta\eta_i}
; \qquad i=1,\dots n,
\label{GL}
\end{equation}
where $D_{c_i}$ and $L_{\eta_i}$ are positive mobilities and may depend on local state variables and interface orientation. The primary distinctions among phase-field models used across different applications arise from the choice of conserved and non-conserved order parameters employed to represent the physical system, the formulation of the various contributions to the total free energy, and the manner in which material properties are interpolated across diffuse interfaces. Microstructure evolution in phase-field modeling is fundamentally driven by the reduction of the system’s total free energy. Accordingly, the general form of the free energy functional in Eq.~(\ref{FreeE}) constitutes a key advantage of the phase-field approach, as it enables the consistent coupling of a wide range of physical phenomena within a unified theoretical framework. In this work, we employ the phase-field method to simulate microstructure evolution governed by grain growth and spinodal decomposition, respectively.

\textit{Grain Growth:} A multi-order-parameter phase-field model \cite{NMoelans} is employed to simulate grain growth in a 2D polycrystalline microstructure. In this framework, a set of non-conserved order parameters $\eta_i$, $i=1, \cdots, n$, is introduced to represent $n$ distinct grain orientations. The total free energy of the system is expressed as a functional of the order parameters and their gradients:
\begin{equation} 
F(\eta_i)  =\int_\Omega \left [f_0(\eta_i) +
\sum_{i=1}^{n}\frac{1}{2}\kappa_i(\nabla \eta_i)^2 \right]dv,
\label{frenergy}
\end{equation}  
where $\kappa_i$ denotes the gradient energy coefficient associated with $\eta_i$, and $f_0$ is the local free energy density, defined as
\begin{equation}
 f_0 = m\left [\sum_{i=1}^n (-\frac{\eta_i^2}{2}+\frac{\eta_i^4}{4})  + \sum_{i=1}^n\sum_{j\not=i}^n \frac{3}{2}\eta_i^2\eta_j^2 + \frac{1}{4}\right ],
\label{ldensity} 
\end{equation}
where $f_0$ yields $n$ equivalent minima with equal depth located at
$(\eta_1, \eta_2,..., \eta_n)=(1,0,...,0), $\\$((0,1,...,0),...,(0,0,...,1))$, corresponding to the stable grain orientations. The temporal evolution of the order parameters is governed by the AC equation:
\begin{equation}
\frac{\partial \eta_i}{\partial
t}=-L\frac{\delta F} {\delta\eta_i}
; \qquad i=1,\dots n.
\label{GLG}
\end{equation}
where $L$ is a constant mobility. While the structure of Eq. (\ref{GLG}) is similar to that of Eq. (\ref{GL}), the free energy functional $F$ differs due to the multi–order-parameter formulation used to represent polycrystalline microstructures.

 \textit{Spinodal  Decomposition:} The regular solution model is used to describe the homogeneous free energy density for the system undergoing spinodal decomposition:
\begin{equation}
 f_{chem}(c) = RT[c\ln c +(1-c)\ln(1-c)] +  \omega c(1-c),
\label{ldensityS} 
\end{equation}
where $c$ denotes the molar fraction of one species in a binary system, $R$ is the gas constant, and $T$ is the absolute temperature, and $\omega$ is the regular solution interaction parameter. A positive value of $\omega$ is chosen to promote phase separation. 
The evolution of the concentration field is governed by the CH diffusion equation
\begin{equation}  
\frac{\partial c}{\partial
t} = {\bf\nabla}\left\{Mc(1-c){\bf\nabla}\left [\frac{\delta f_{chem}} {\delta c}-\epsilon{\bf\nabla}^2 c\right ]\right\},
\label{CHS}
\end{equation} 
where $M$ is the mobility and $\epsilon$ is the gradient energy coefficient controlling interfacial width. Equations (\ref{ldensityS}) and (\ref{CHS}) are solved subject to no-flux boundary conditions.

\section{Fully Convolutional Spatiotemporal Learning Framework for Microstructure Evolution Prediction}
\label{sec:DLM}
\begin{figure}
\includegraphics[width=\linewidth, trim=0cm 6cm 0cm 6cm]{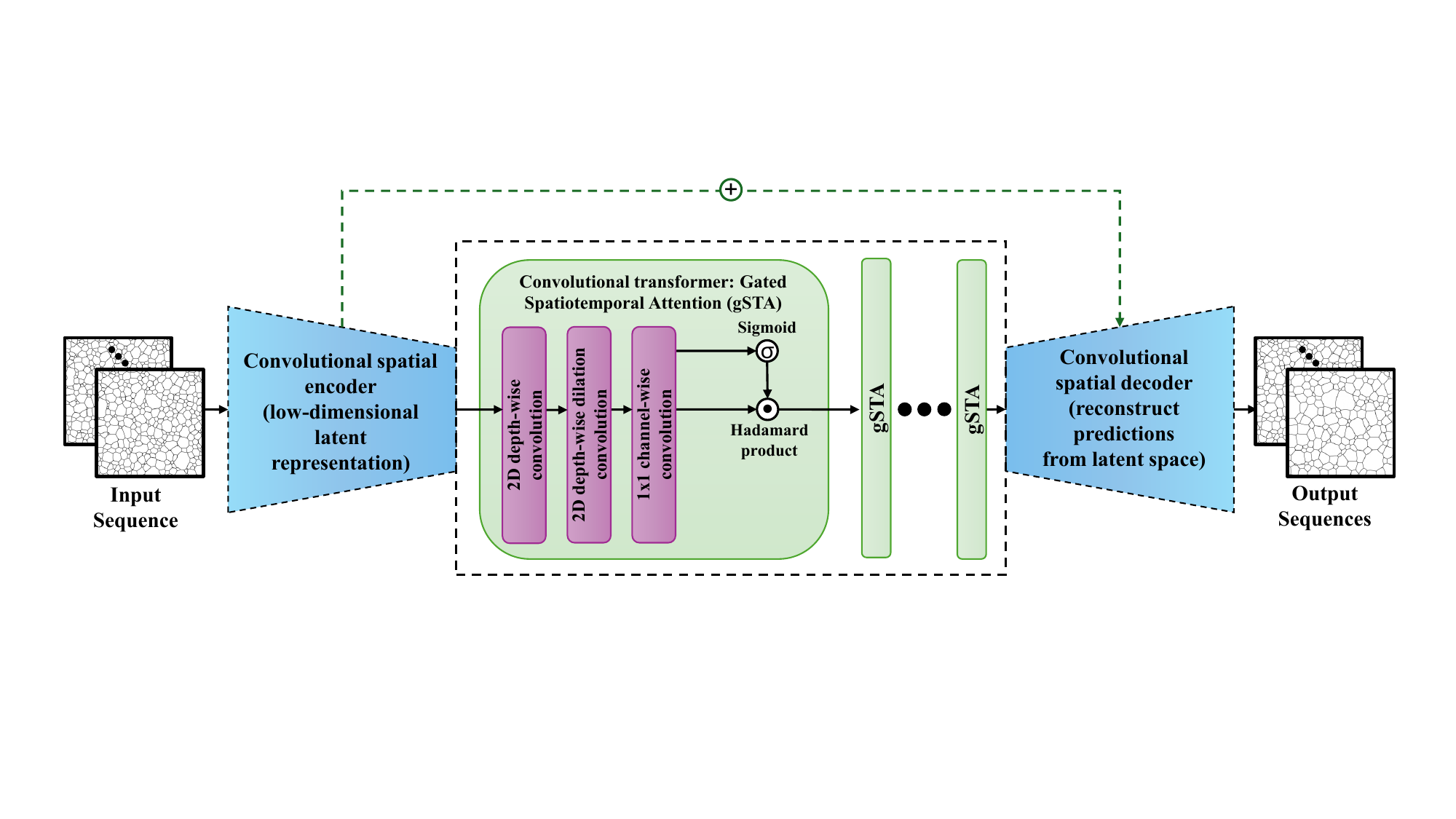}
\caption{Flowchart of the proposed fully convolutional spatiotemporal framework for microstructure evolution prediction. An input sequence of length $T$ is first processed by a \textit{convolutional} spatial encoder to extract microstructural spatial correlations and project the data into a compact latent feature space. The latent representations are then passed through stacked Gated Spatiotemporal Attention (gSTA) modules, which serve as the temporal translator of the network and model the dynamic evolution patterns of the microstructure. Within this module, the encoded features are propagated and transformed to generate forecast frames corresponding to future time steps $T+1$ through $T+T'$. Finally, a \textit{convolutional} spatial decoder reconstructs the predicted microstructure fields by upsampling the latent representations back to the original spatial resolution, producing the output sequence.}
\label{SimVP}
\end{figure}

The phase-field formulations described in Section~\ref{sec:PFM} provide a physically rigorous framework for modeling microstructure evolution driven by grain growth and spinodal decomposition. While these models offer high-fidelity representations of microstructural dynamics, their numerical solution requires repeated integration of nonlinear PDEs over fine spatial and temporal resolutions, leading to significant computational cost. 
In this section, we introduce a fully convolutional spatiotemporal learning framework for data-driven prediction of microstructure evolution. The proposed approach learns the evolution operator directly from simulation data, mapping a sequence of past microstructure states to future configurations. By formulating microstructure evolution as a spatiotemporal forecasting problem, the framework enables efficient long-horizon prediction while maintaining scalability to high-resolution microstructural fields.

\subsection{Problem Formulation}
Given a temporal sequence of microstructure images, $\mathcal{X}^{1:T} = \{x^1, x^2, \dots, x^T\}$, our objective is to predict the subsequent $T'$ future microstructure dynamics $\mathcal{Y}^{1:T'} = \{x^{T+1}, \dots, x^{T+T'}\}$. Each image $x^t \in \mathbb{R}^{H \times W}$ represents a snapshot of the microstructure at time $t$, where $H$ and $W$ denote the spatial resolution of the computational domain. These images correspond to discretized phase-field variables, such as concentration fields or order parameters, generated by phase-field simulations.

The learning task is to construct a predictive mapping $F_\Theta: \mathcal{X}^{1:T} \rightarrow \mathcal{Y}^{1:T'}$, parameterized by a neural network with learnable weights $\Theta$, that approximates the underlying spatiotemporal evolution of the microstructure. The optimal parameters $\Theta^*$ are obtained by minimizing a loss function $\mathcal{L}$ that measures the discrepancy between predicted and ground-truth future sequences: 
\begin{equation}
    \Theta^* = \arg \min_\Theta \mathcal{L}\left(F_\Theta(\mathcal{X}^{1:T}), \mathcal{Y}^{1:T'}\right).
\end{equation}
This formulation explicitly captures both the spatial structure and temporal evolution of microstructures, and can be interpreted as a data-driven approximation of the phase-field evolution operator. By learning this mapping directly from simulation data, the proposed framework enables efficient multi-step prediction of microstructure dynamics without explicitly solving the governing partial differential equations.

\subsection{Fully Convolutional Spatiotemporal Network Architecture}

In this work, we adopt SimVPv2 \cite{tan2022simvpv2} as the backbone model for microstructure evolution prediction. In contrast to traditional video prediction models that rely on encoder–decoder frameworks with recurrent structures, SimVPv2 is a fully convolutional architecture that decomposes spatiotemporal learning into three streamlined modules: a \textit{spatial encoder}, a \textit{temporal translator}, and a \textit{spatial decoder}, all implemented using lightweight convolutional blocks, as illustrated in Fig.~\ref{SimVP}. This design significantly simplifies the modeling pipeline while maintaining strong spatiotemporal representation capability. In particular, the architecture employs depth-wise spatial convolutions, dilated convolutions, and $1 \times 1$ channel-wise convolutions to achieve efficient yet physically meaningful spatiotemporal representation, which is especially suitable for diffusion-driven microstructure evolution processes.

Traditional autoencoders primarily focus on single-image reconstruction by learning an encoding–decoding pipeline that minimizes reconstruction error. The proposed spatiotemporal learning model extends this paradigm to the temporal domain by encoding a sequence of input microstructure states $\mathcal{X}^{1:T}$ and directly predicting future evolutions $\mathcal{Y}^{1:T'}$, thereby jointly modeling spatial structures and their temporal dynamics. Specifically, the spatial encoder extracts rich feature representations from each input frame, capturing essential microstructural characteristics such as grain boundaries, texture patterns, and phase distributions. The encoder is constructed using 2D depth-wise convolutions, which perform channel-wise spatial filtering in the latent space. This design allows each latent feature channel—encoding distinct morphological descriptors such as interface gradients or local phase contrast—to be refined independently before cross-channel interaction, thereby preserving physically meaningful spatial structures while reducing parameter redundancy and improving computational efficiency. These spatial features are then forwarded to the temporal translator, which is responsible for learning how such spatial patterns evolve over time. Finally, the spatial decoder upsamples the latent spatiotemporal features to generate future microstructure fields at the desired spatial resolution.

A key component of the proposed model is the gated Spatiotemporal Attention (gSTA) module employed within the temporal translator. In contrast to conventional temporal convolutions, which apply fixed, translation-invariant kernels across space and time, gSTA introduces a learnable attention-based gating mechanism that adaptively modulates spatiotemporal feature interactions according to evolving microstructural patterns. Operating on latent feature tensors $\mathbf{Z} \in \mathbb{R}^{\tilde{T} \times \tilde{C} \times \tilde{H} \times \tilde{W}}$, gSTA jointly captures temporal progression and spatial heterogeneity arising from interface motion, grain coarsening, and phase separation dynamics. Here, $\tilde{T}$ denotes the temporal length of the encoded latent sequence, $\tilde{C}$ is the number of latent feature channels, and $\tilde{H}$ and $\tilde{W}$ are the spatial dimensions after encoding.  Concretely, gSTA constructs query, key, and value representations using parallel convolutional operators: 
$
\mathbf{Q} = \phi_Q(\mathbf{Z}), 
\mathbf{K} = \phi_K(\mathbf{Z}), 
\mathbf{V} = \phi_V(\mathbf{Z}),
$
where $\phi_Q$, $\phi_K$, and $\phi_V$ denote lightweight convolutional operators that preserve local spatial structure associated with grain boundaries and phase interfaces. These operators are implemented using 2D depth-wise convolutions combined with depth-wise dilated convolutions to enlarge the receptive field without excessive downsampling. The dilation mechanism enables the model to capture long-range spatial dependencies induced by diffusion and curvature-driven interface motion, which are fundamental to coarsening and phase separation dynamics, while maintaining linear computational complexity. Subsequently, $1 \times 1$ channel-wise convolutions are employed to mix information across feature channels, facilitating nonlinear coupling between latent physical descriptors and enabling efficient feature fusion analogous to interactions among multiple driving forces governing microstructure evolution.

A causal, localized spatiotemporal aggregation is then performed over a neighborhood $\mathcal{N}(t,\mathbf{p})$ centered at time index $t$ and spatial location $\mathbf{p}$, with the causality constraint $s \le t$ enforcing physically consistent temporal evolution. The attention weights are defined as
$
\alpha_{t,\mathbf{p}}(s,\mathbf{q})
=
\operatorname{softmax}_{(s,\mathbf{q}) \in \mathcal{N}(t,\mathbf{p})}\\
\left(
\frac{\langle \mathbf{Q}_{t,\mathbf{p}}, \mathbf{K}_{s,\mathbf{q}} \rangle}{\sqrt{d}}
\right),
$
where $d$ denotes the embedding dimension used for attention computation, typically chosen such that $d \le \tilde{C}$ to balance expressiveness and computational efficiency, and the aggregated spatiotemporal feature is given by
$
\widetilde{\mathbf{V}}_{t,\mathbf{p}}
=
\sum_{(s,\mathbf{q}) \in \mathcal{N}(t,\mathbf{p})}
\alpha_{t,\mathbf{p}}(s,\mathbf{q}) \, \mathbf{V}_{s,\mathbf{q}} .
$
This localized attention mechanism enables the model to focus on regions where microstructural changes are most pronounced, such as rapidly migrating grain boundaries or unstable phase interfaces, while suppressing interactions in relatively static regions. To further regulate information flow, gSTA introduces a learnable gating mechanism,
$
\mathbf{G}
=
\sigma \left( \psi \big( [\mathbf{Z}, \widetilde{\mathbf{V}}] \big) \right), 
\mathbf{Z}' = \mathbf{Z} + \mathbf{G} \odot \widetilde{\mathbf{V}},
$
where $[\cdot,\cdot]$ denotes channel-wise concatenation, $\psi(\cdot)$ is a lightweight convolutional block, $\sigma(\cdot)$ is the sigmoid activation,  $\odot$ denotes element-wise multiplication, and $\mathbf{Z}'$ is the updated latent spatiotemporal feature tensor. This gated residual update selectively emphasizes regions undergoing active microstructure evolution, such as migrating grain boundaries or phase interfaces, while preserving relatively stable grain interiors.

Importantly, gSTA is implemented using localized convolutional attention rather than global self-attention, resulting in linear computational complexity with respect to spatial resolution. Moreover, the strictly causal temporal structure prevents information leakage from future states, ensuring physically consistent prediction of microstructure evolution. From a modeling perspective, the fully convolutional and non-recurrent formulation avoids hidden-state accumulation, improving training stability and mitigating long-term error propagation during long-horizon prediction of evolving microstructures.

\subsection{Training Strategy and Loss Function}
The proposed fully convolutional spatiotemporal framework is trained in a self-supervised manner using sequences of microstructure images generated from the phase-field simulations described in Section~2. Each training sample consists of an input sequence of $\tilde{T}$ consecutive microstructure states and a corresponding target sequence of length $T'$, following the formulation introduced in Section~3.1. No additional labels or physical parameters are required during training, as the network learns the evolution dynamics directly from spatiotemporal data.

During training, the network takes the input sequence $\mathcal{X}^{1:\tilde{T}}$ and predicts the future microstructure sequence $\hat{\mathcal{Y}}^{1:T'}$ in a single forward pass. In contrast to autoregressive recurrent models that generate predictions sequentially and propagate hidden states through time, the fully convolutional architecture enables parallel prediction across all future time steps, resulting in improved computational efficiency and training stability for long-horizon forecasting.

To supervise learning, we employ a mean-squared error (MSE) loss defined over the predicted and ground-truth microstructure fields:
\begin{equation}
\mathcal{L}
=
\frac{1}{T'} \sum_{t=1}^{T'}
\left\|
\hat{x}^{\tilde{T}+t} - x^{\tilde{T}+t}
\right\|_2^2 ,
\end{equation}
where $\hat{x}^{\tilde{T}+t}$ and $x^{\tilde{T}+t}$ denote the predicted and reference microstructure fields at future time step $t$, respectively. This loss directly penalizes discrepancies in spatial field values and encourages accurate reconstruction of evolving microstructural patterns.

For microstructure evolution problems, accurate long-term prediction is essential, as small local errors can accumulate and lead to qualitatively incorrect morphologies at later times. The use of parallel multi-step supervision encourages the model to learn stable long-range evolution behavior rather than relying solely on short-term temporal correlations. In practice, this training strategy improves robustness when predicting long-horizon microstructure dynamics, such as grain growth and spinodal decomposition.

\section{Experimental Setting}
\label{sec:exp}

\subsection{Phase-Field Simulation Datasets}

The training and evaluation data consist of previously published phase-field simulation datasets \cite{yang2021self}, which provide ground-truth microstructure evolution for grain growth and spinodal decomposition. To verify robustness and reproducibility, additional simulations are performed under identical physical and numerical settings. No modifications to the governing equations, model parameters, or numerical solvers are introduced in this work. Phase-field methods provide a diffuse-interface framework for modeling microstructure evolution. In these models, grain orientations or phase concentrations are represented by continuous order parameters whose dynamics are governed by thermodynamically derived PDEs. This formulation naturally captures interfacial motion, topological transitions, and long-time morphological evolution without explicit interface tracking.

\paragraph{Grain Growth}
Grain growth is simulated using a multi–order-parameter phase-field model, where the evolution of grain orientations is governed by the AC-type equations described in Section~2. In all simulations, the dimensionless parameters are set to $N = 100$, $m = 1$, $\nu = 1$, and $L = 1$. The initial polycrystalline microstructure is generated using Voronoi tessellation with 100 randomly distributed grains. Single-channel images of the polycrystalline structure are constructed by assigning pixel values according to
$
\sum_{i=1}^{N} \eta_i^3,
$
where $\eta_i$ denotes the order parameter associated with the $i$th grain. This representation yields pixel values close to zero in grain boundary regions and values close to one within grain interiors. 
The governing equations are solved using a forward Euler finite-difference scheme with periodic boundary conditions. The spatial grid spacing is set to $\Delta x = \Delta y = 1$, and the time-step size is chosen as $\Delta t = 0.2$. Microstructure snapshots are recorded at regular intervals to form spatiotemporal sequences for training and evaluation.

\paragraph{Spinodal Decomposition}
Spinodal decomposition is simulated using the CH formulation described in Section~2. The dimensionless parameters are set to $\omega = 0.27397$, $\epsilon = 0.1682$, and mobility $M = 1$, with spatial grid spacing $\Delta x = \Delta y = 1$ in all simulations. The governing equations are solved using an implicit variable-order backward differentiation formula (BDF) solver implemented in COMSOL Multiphysics \cite{multiphysics1998introduction}. An average dimensionless time-step size of $4.01$ is used. 
Simulation outputs are recorded at a temporal interval of 1{,}500 solver steps, corresponding to approximately 370 solver iterations between consecutive frames. The resulting concentration fields are stored as image sequences representing the temporal evolution of phase separation.

\subsection{Dataset Construction and Preprocessing}
\label{sec:DataPre}

All microstructure sequences are organized as tensors of shape
$
\mathbb{R}^{B \times T \times C \times H \times W},
$
where $B$ denotes the number of spatiotemporal sequences (samples), $T$ is the number of frames in each sequence, $C$ is the number of image channels, and $H$ and $W$ represent the spatial resolution. In this work, all microstructure images are represented as single-channel grayscale images ($C=1$). 
For each training sample, we use $\tilde{T}=10$ consecutive frames as input and predict the subsequent $T'=90$ frames, resulting in a total temporal window of 100 frames per sample. To construct the dataset, a sliding window is applied to each simulation sequence. Given a stride size $S$, this procedure yields approximately
$
B \times \left( \left\lfloor \frac{T - 100}{S} \right\rfloor + 1 \right)
$
training or evaluation samples, where overlapping windows allow efficient reuse of long simulation trajectories. 

The proposed framework supports variable-length input sequences up to $\tilde{T}=10$ frames. When fewer than 10 input frames are available, zero-padding is applied at the beginning of the input sequence to maintain a fixed input length. For example, a sequence with 5 available frames is padded with 5 leading zero frames before being passed to the model. This padding strategy enables flexible use of pre-trained models without modifying the network architecture. 
To enable prediction beyond the nominal horizon of $T'=90$ frames, an iterative inference strategy is adopted: the final 10 predicted frames are fed back into the model as input to generate subsequent predictions. This process can be repeated to produce microstructure evolution forecasts of arbitrary temporal length.

All datasets are partitioned into disjoint training, validation, and test sets at the trajectory level to avoid temporal information leakage. 
For grain growth, we generated approximately 1,070 simulation trajectories, each consisting of 200 frames at a spatial resolution of $64 \times 64$. From these trajectories, we extracted 11,000 training clips out of 1000 and 750 validation clips out of 70 using a sliding-window procedure with window length 10. 
The test set comprises 100 independent testing trajectories, each with 200 frames at $256 \times 256$ resolution. For evaluation, each 200-frame trajectory is further segmented into 100-frame clips, yielding 1,100 testing sequences in total.

For spinodal decomposition, the training set contains approximately 580 trajectories and the validation set contains 40 trajectories, both at $64 \times 64$ resolution. Applying the same window length 10 sliding-window procedure yields 6,380 training clips and 440 validation clips. The test set contains 50 independent trajectories, each with 210 frames at $256 \times 256$ resolution. For evaluation, each trajectory is segmented into 100-frame clips, resulting in the corresponding set of testing sequences.

\subsection{Fully Convolutional Spatiotemporal Model Training and Inference}

We employ 4 spatial attention blocks with 64 hidden channels for the spatial encoder and decoder. The temporal translator consists of 8 temporal attention blocks with 256 hidden channels. Greater capacity is allocated to the temporal module because long-horizon temporal prediction is more challenging than spatial reconstruction and requires modeling accumulated nonlinear evolution dynamics. Each model is trained for up to 200 epochs with batch size 1. The average training time per epoch is approximately 1 hour and 28 minutes for the grain growth dataset and 26 minutes for the spinodal decomposition dataset. After training on $64 \times 64$ resolution data, the fully convolutional models are directly evaluated on $256 \times 256$ simulations without retraining. Owing to the resolution-agnostic design of the architecture, no modification to the network structure is required for higher-resolution inference. 
All experiments were performed on the AI Panther system, equipped with A100 SXM4 GPUs, at the Florida Institute of Technology.

\subsection{Performance Metrics}
The accuracy of predicted microstructure evolution via deep learning can be assessed by the pixel-wise comparison between ground truth and prediction. In order to quantitatively compare between predictions and ground truth in microstructure evolution, we will use the following statistical measurements, root-mean-square error (RMSE) and structure similarity index measure (SSIM) \cite{ZWang04} to the microstructures from both predictions and ground truth.  The RMSE and SSIM are evaluated every 5 frames of each prediction result. RMSE and SSIM are defined as follows:

\begin{equation}
\text{RMSE} = \sqrt{ \sum_{i=1}^{N_{x}} \sum_{j=1}^{N_{y}}\frac{1}{N_x}\frac{1}{N_y}  \left( y_{g}(i,j) - y_{p}(i,j) \right)^2 },
\label{RMSE}
\end{equation}
where subscripts g and p represent ground truth and prediction data respectively, $N_{x}$ is the number of pixel rows in the data, $N_{y}$ is the number of pixel columns in the data, and $y(i,j)$ is the pixel value at a specific coordinate.

\begin{equation}
\text{SSIM}(g, p) = \frac{ (2 \mu_g \mu_p + C_1)(2 \sigma_{gp} + C_2) }{ (\mu_g^2 + \mu_p^2 + C_1)(\sigma_g^2 + \sigma_p^2 + C_2) },
\label{SSIM}
\end{equation}
where subscripts g and p represent ground truth and prediction data respectively, $\mu$ is the pixel sample mean, $\sigma$ is the variance, $\sigma_{gp}$ is the covariance, and $C_1 = 0.01$ and $C_2 = 0.03$ represent constants.

In addition to pixel-level and structural metrics, for the grain growth problem we evaluate statistical morphology consistency by computing the grain size distribution (GSD). The predicted GSD is compared with that of the ground truth to assess whether the model preserves physically meaningful coarsening behavior beyond pixel-wise similarity. For visualization consistency across different models and time steps, all predicted frames are displayed using a fixed intensity range of $[0, 0.6]$. This clipping is applied only for visualization and does not affect quantitative metric computation.

\section{Numerical Results}
\label{sec:res}
 
In this section, we evaluate the proposed fully convolutional spatiotemporal framework on grain growth and spinodal decomposition. The experiments are designed to assess both short-term forecasting accuracy and long-horizon extrapolation capability, as well as the preservation of physics-based statistical properties. We first present results for grain growth, followed by spinodal decomposition.

\begin{figure}
\centering
\includegraphics[width=\linewidth]{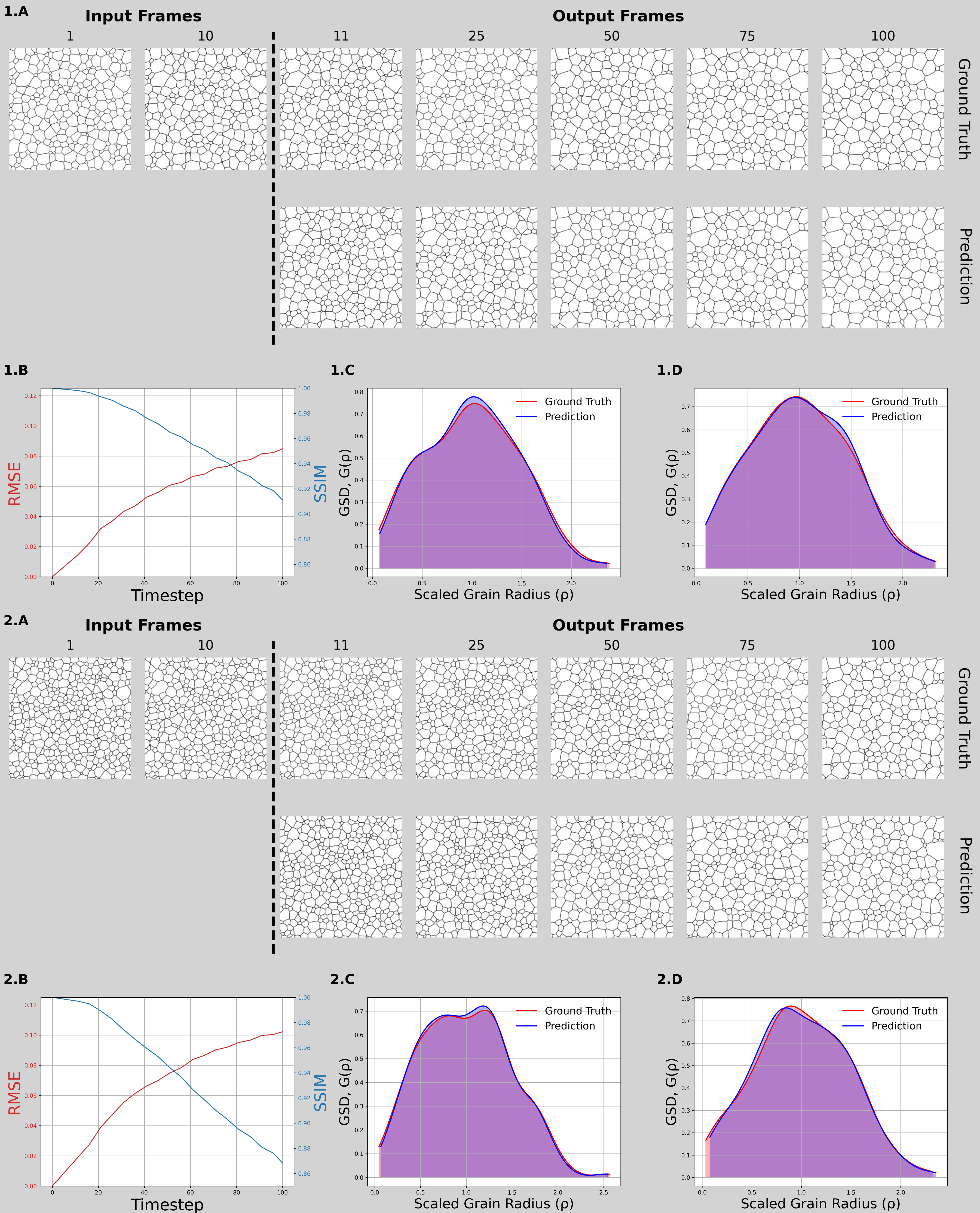}
\caption{\textbf{Grain growth prediction (10 input frames, 90 output frames) using the proposed fully convolutional spatiotemporal model:} Two representative test samples are shown (indexed numerically). (A) Predicted and ground-truth microstructure frames at $t=11, 25, 50, 75,$ and $100$; (B) RMSE and SSIM evaluated every 5 frames over the prediction horizon; (C--D) Grain-size distributions (GSDs) for prediction and ground truth at $t=25$ and $t=100$, respectively.}
\label{fig2}
\end{figure}

\subsection{Grain Growth}
The microstructural evolution in the process of grain growth described by grain size and GSD has a significant influence on the mechanical, thermal,
and electrical properties of engineering materials. During the process of grain growth  the positions of grain boundaries are changing while the volume is transferred from one grain to another. Theories of grain growth fall into two broad classes: those rooted in topology and those rooted in stochastic behavior. The fundamental equation of the topological models is the von Neumann-Mullins rule for 2D grain growth. As long as the grain boundary energies are isotropic, the dihedral
angles where the three grain boundaries meet must all be $120^{\circ}$. Therefore, grains with more than six sides have concave boundaries and will grow, and grains with fewer than six sides will have convex boundaries and shrink. Accordingly, during 2D ideal grain growth, all individual grains should satisfy the von Neumann-Mullins equation \cite{WWMullins56}:
\begin{equation}
\frac{dA}{dt} = M_g\gamma\frac{\pi}{3}(N-6),
\label{GGk}
\end{equation}
where $M_g$, $A$, and $N$ are the boundary mobility, the grain area, and the number of sides, respectively. Researchers have sought a similar rule for 3D grain growth.

Hillert \cite{MHillert65} proposed that there is a characteristic grain size above which all grains grow and below which all grains shrink. On this basis, he developed a model that yields a parabolic growth law, which can be expressed as follows:

\begin{equation}
<R(t)>^2 -<R(t_0)>^2 = m_R t
\label{GGRk}
\end{equation}
where  $t_0$ is an initial time; $<R(t)>$ is
the average grain radius at the time of $t$; $m_R$ is the proportional
constant.

According to Hillert's mean field theory, the GSD is expressed as follows:

\begin{equation}  G_H(\rho)= \left\{\!\!
\begin{array}{cc} [(2e)^{\lambda}\frac{\lambda\rho}{(2-\rho)^{2+\lambda}}]
\exp [\frac{-2\lambda\rho}{(2-\rho)}] &  (\rho<2),\\ \\ 0 &  (\rho \ge 2).
\end{array}\right. 
\label{PSDGG}
\end{equation} 
where $\rho = R/<R>$, $\lambda =  2 $ in 2D system, and $\lambda = 3$ in 3D  system. On
the basis of Hillert's theory, This distribution is identical (except for a scale factor) to that resulting from Wagner's coarsening with surface reaction-limited kinetics. It is less symmetric than the observed GSDs and predicts a much smaller maximum grain size.

Using the trained model, we perform three evaluation scenarios on grain growth simulations at $256 \times 256$ resolution. The first one matches the training task by predicting 90 future frames from 10 input frames (10$\rightarrow$90), and primarily tests resolution generalization. The second scenario extends the forecasting horizon to 190 future frames from 10 prior frames (10$\rightarrow$190) using iterative roll-out, where the last 10 predicted frames are recursively used as the next input. The third one evaluates robustness under limited observations by predicting 95 future frames from only 5 prior frames (5$\rightarrow$95), in which we pad the input with 5 leading zero frames.

We evaluate prediction quality using RMSE and SSIM (Eqs.~\ref{RMSE}--\ref{SSIM}) computed every 5 frames. Since grain growth is governed by curvature-driven coarsening and exhibits characteristic statistical behavior, RMSE/SSIM alone may not fully reflect physical fidelity. We therefore additionally compute the GSD from the predicted sequences and compare it with the ground truth, with Eq.~\ref{PSDGG} serving as a reference for expected coarsening statistics. This combined evaluation assesses both pixel-level accuracy and physics-consistent morphology evolution.

\subsubsection{Predictions of 90 future frames from 10 input frames (10$\rightarrow$90)}

For the 10$\rightarrow$90 task, 100 test samples are evaluated using the trained model. Two representative examples are shown in Fig.~\ref{fig2}. For compact visualization, the input sequence is displayed at $t=1$ and $t=10$, while the predicted and ground-truth outputs are shown at $t=10, 25, 50, 75,$ and $100$ (Figs.~\ref{fig2}.1.A and \ref{fig2}.2.A). The predictions closely match the ground truth, especially at early and intermediate times, while small discrepancies become more visible at later time steps.
Figs.~\ref{fig2}.1.B and \ref{fig2}.2.B report the RMSE and SSIM, computed every 5 frames across the prediction horizon. Both examples maintain low pixel-wise error, with RMSE remaining below 0.11 and SSIM above 0.86 throughout the forecast window, indicating stable prediction accuracy. To assess morphology fidelity beyond pixel-wise agreement, Figs.~\ref{fig2}.1.C--D and \ref{fig2}.2.C--D compare the GSDs of predictions and ground truth at $t=25$ and $t=100$. The predicted GSDs closely follow the ground-truth distributions at both time points, suggesting that the model preserves the coarsening-driven statistical evolution of grain sizes without explicit access to the underlying governing equations.

\begin{figure}
\centering
\includegraphics[width=\linewidth]{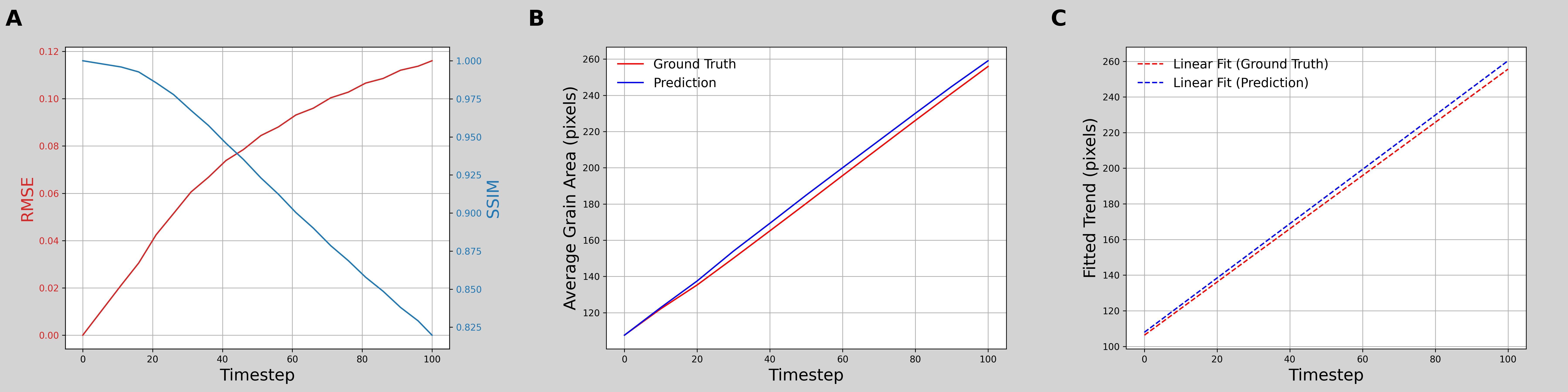}
\caption{\textbf{Accuracy metrics on grain growth prediction (10 input frames, 90 output frames):} (A) Dataset-averaged RMSE and SSIM evaluated every 5 frames over the prediction horizon; (B) Temporal evolution of the mean grain area for predicted and ground-truth sequences (computed every 10 frames); (C) Corresponding linear regression fits highlighting agreement in coarsening trends between prediction and ground truth.}
\label{fig3}
\end{figure}

Aggregate results over all 100 test samples are summarized in Fig.~\ref{fig3}. Fig.~\ref{fig3}.A plots the mean RMSE and SSIM across the dataset, computed every 5 frames, demonstrating consistently strong average performance. Fig.~\ref{fig3}.B compares the evolution of the mean grain area (computed every 10 frames) between predicted and ground-truth sequences, and Fig.~\ref{fig3}.C shows the corresponding linear trend in mean grain area over time. The close agreement in these statistics suggests that the model captures not only visual similarity but also the characteristic coarsening behavior of the grain growth process.

Notably, the model is trained only on $64\times64$ clips but is directly evaluated on $256\times256$ simulations without retraining or architectural modification. The qualitative agreement in the predicted microstructures, together with consistently low RMSE / high SSIM  and closely matched GSDs at both intermediate and late times, confirms the promising generalization capability of the fully convolutional framework to substantially higher spatial resolution.

\begin{figure}
\centering
\includegraphics[width=0.95\linewidth]{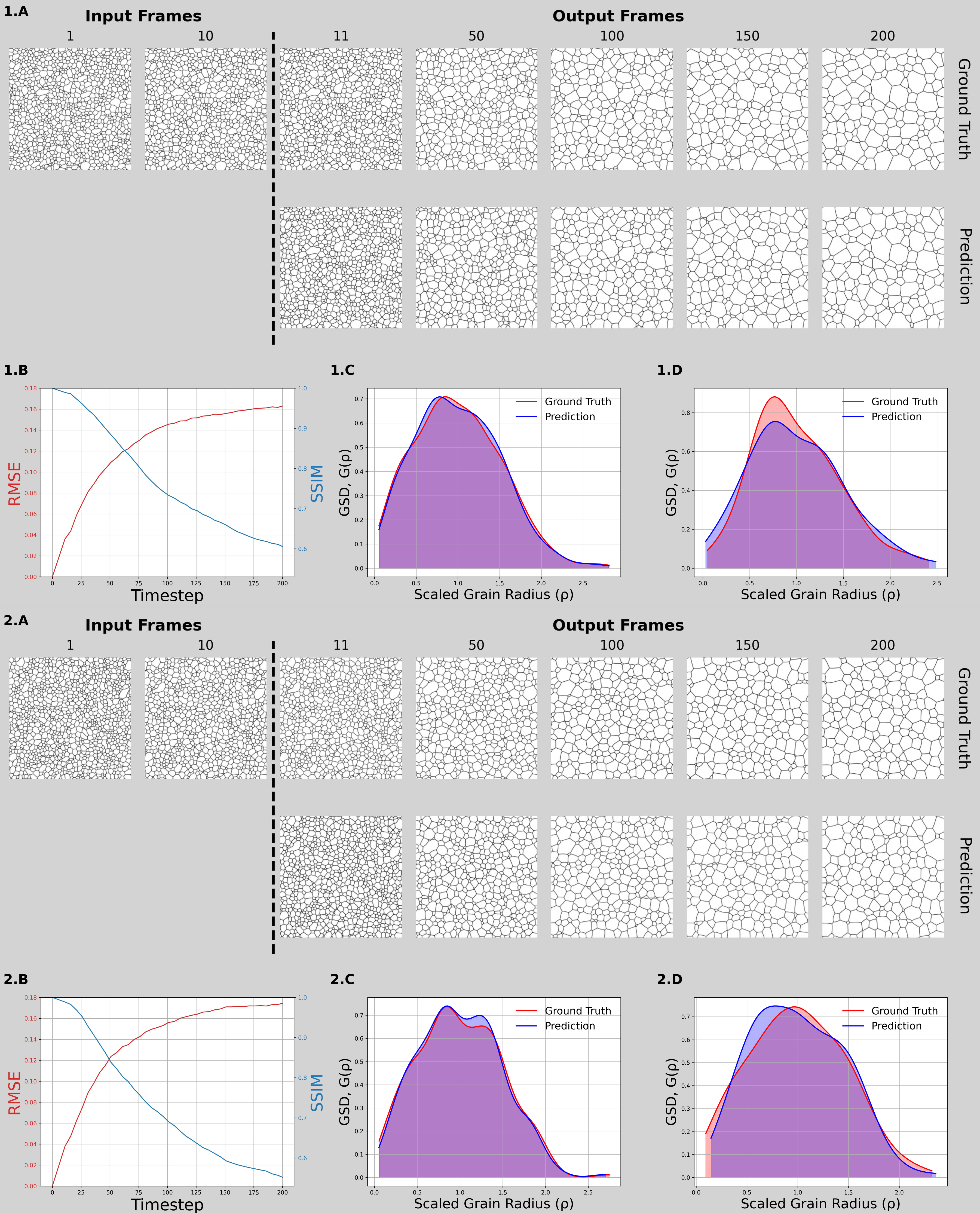}

\caption{\textbf{Grain Growth prediction (10 input frames, 190 output frames) using the proposed fully convolutional spatiotemporal model:} Two representative test samples are shown (indexed numerically). (A) Predicted and ground-truth microstructure frames at t = 11, 50, 100, 150, 200; (B) RMSE and SSIM evaluated every 5 frames over the prediction horizon; (C--D) Grain size distributions (GSD) for prediction and ground truth at t=50 and t=200, respectively.}
\label{fig4}
\end{figure}

\subsubsection{Predictions of 190 future frames from 10 input frames (10$\rightarrow$190)}

For the 10$\rightarrow$190 task, we evaluate the trained model on 100 test samples using iterative roll-out to extend the prediction horizon beyond the nominal 90-frame output length. Two representative examples are shown in Fig.~\ref{fig4}. For compact visualization, the input sequence is displayed at $t=1$ and $t=10$, while the predicted and ground-truth outputs are shown at $t=11, 50, 100, 150,$ and $200$ (Figs.~\ref{fig4}.1.A and \ref{fig4}.2.A). The prediction frames demonstrate higher accuracy in early stages of prediction, with errors increasing at later timesteps. Despite this degradation, the model preserves grain interface structures throughout the extended extrapolation task, noticeably in the center regions of its prediction frames. Figs.~\ref{fig4}.1.B and \ref{fig4}.2.B present the RMSE and SSIM metrics, measured at 5-frame intervals, where RMSE remains below 0.18 and SSIM above 0.5 for both samples. Figs.~\ref{fig4}.1.C--D and \ref{fig4}.2.C--D display the GSD statistics at timesteps t=50 and t=200. The results between the predicted and ground-truth sequences are closely-distributed at timestep 50. By timestep 200 the predictions diverge from the ground truth; however, the overall spread of the distributions remains comparable. These results suggest that the network captures the underlying grain size statistics even when forecasting beyond its trained prediction horizon.

\begin{figure}
\centering
\includegraphics[width=\linewidth]{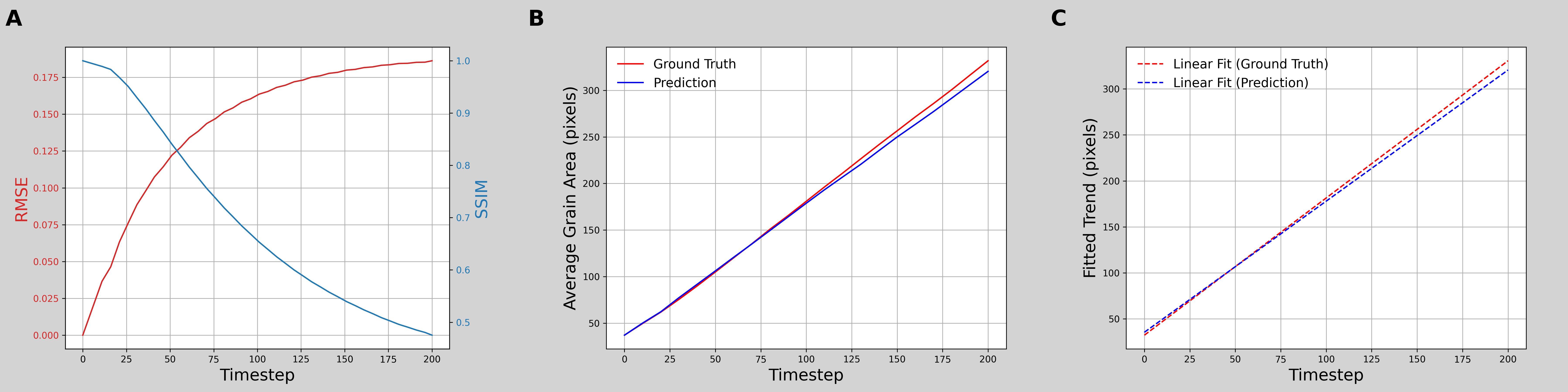}
\caption{\textbf{Accuracy metrics on grain growth prediction (10 input frames, 190 output frames):} (A) Dataset-averaged RMSE and SSIM evaluated every 5 frames over the prediction horizon; (B) Temporal evolution of the mean grain area for predicted and ground-truth sequences (computed every 10 frames); (C) Corresponding linear regression fits highlighting agreement in coarsening trends between prediction and ground truth.}
\label{fig5}
\end{figure}

Fig.~\ref{fig5} summarizes the average evaluation metrics across the 100 test samples for this prediction task. Fig.~\ref{fig5}.A shows the mean RMSE and SSIM over the dataset, computed at 5-frame intervals. Although larger errors are expected for long-horizon forecasting, the predicted frames remain free of noticeable visual artifacts and preserve realistic microstructural morphology, as seen in Fig.~\ref{fig4}. Fig.~\ref{fig5}.B compares the evolution of the mean grain area between the predicted and ground-truth sequences, measured every 10 frames, while Fig.~\ref{fig5}.C presents the corresponding linear trends. The close agreement between these trends indicates that the network preserves the underlying physics-based characteristics of grain growth despite degradation in pixel-wise metrics.

\subsubsection{Predictions of 95 future frames from 5 input frames (5$\rightarrow$95)}
\begin{figure}[h!]
\centering
\includegraphics[width=0.95\linewidth]{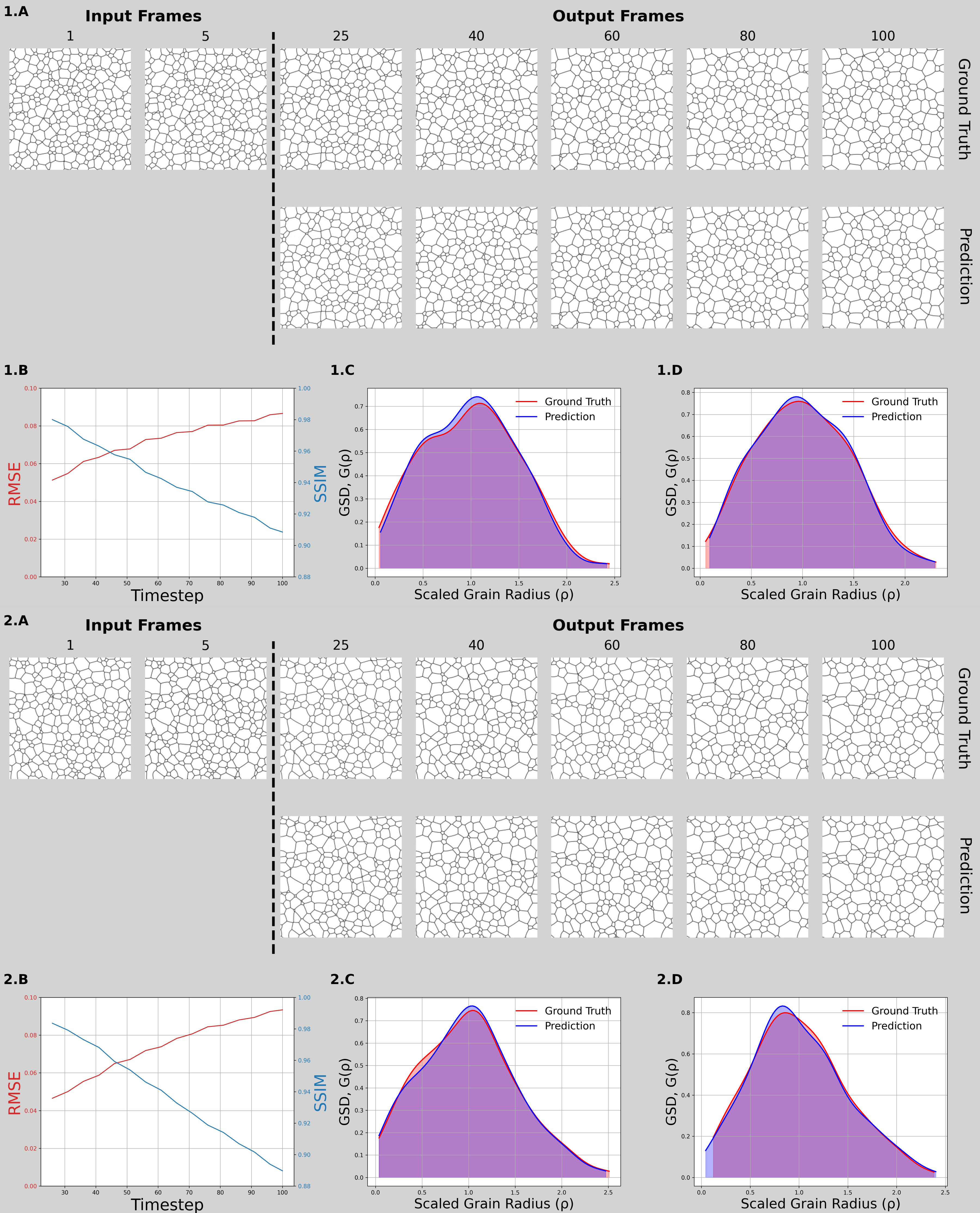}

\caption{\textbf{Grain growth prediction (5 input frames, 95 output frames) using the proposed fully convolu-
tional spatiotemporal model:} Two representative test samples are shown (indexed numerically). (A) Predicted and ground-truth microstructure frames at t = 25, 40, 60, 80, 100; (B) RMSE and SSIM computed between the predicted and ground-truth sequences; (C--D) Grain size distributions evaluated (GSDs) for prediction and ground truth at t=25 and t=100, respectively.}
\label{fig6}
\end{figure}

For the 5$\rightarrow$95 task, we evaluate the model on 110 test samples to assess robustness under reduced temporal context. Two representative examples are shown in Fig.~\ref{fig6}. For compact visualization, the input sequence is displayed at $t=1$ and $t=5$, while the predicted and ground-truth outputs are shown at $t=25, 40, 60, 80,$ and $100$ (Figs.~\ref{fig6}.1.A and \ref{fig6}.2.A). Despite the reduced input context, the predictions closely match the ground truth in both samples, including at later timesteps. While minor differences are observed along domain boundaries, the predicted microstructures strongly resemble the ground-truth morphology. The RMSE and SSIM results of each sample are plotted in Figs.~\ref{fig6}.1.B and \ref{fig6}.2.B, starting from timestep t=25 and measured at 5-frame intervals.  
For both samples, RMSE remains below 0.10 and SSIM exceeds 0.88 across the forecasting horizon, indicating robust predictive performance during the later stages of evolution. In Figs.~\ref{fig6}.1.C--D and \ref{fig6}.2.C--D, the GSD plots are displayed at timesteps t=25 and t=100. The predicted distributions closely align with the ground truth, suggesting that the model preserves the underlying statistical characteristics of grain growth even when provided with limited temporal context.

\begin{figure}[h!]
\includegraphics[width=\linewidth]{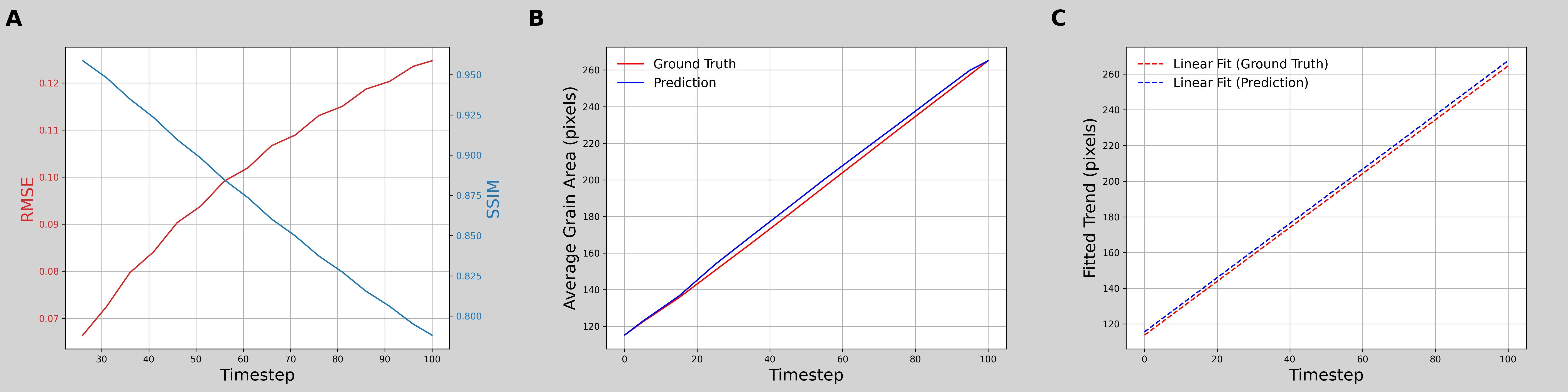}
\caption{\textbf{Accuracy metrics on grain growth prediction (5 input frames, 95 output frames):} (A) Dataset-averaged RMSE and SSIM evaluated every 5 frames over the prediction horizon; (B) Temporal evolution of the mean grain area for predicted and ground-truth sequences (computed every 10 frames); (C) Corresponding linear regression fits highlighting agreement in coarsening trends between prediction and ground truth.}
\label{fig7}
\end{figure}

Fig.~\ref{fig7} shows the averaged result metrics of the 110 test samples. Fig.~\ref{fig7}.A shows the mean RMSE and SSIM over the dataset, starting from timestep t=25 computed at 5-frame intervals. These results indicate that the network maintains strong predictive accuracy despite the reduced temporal context. The mean grain area across all samples is plotted in Fig.~\ref{fig7}.B between the ground truth and predicted sequences. The linear trend of these plots are shown in Fig.~\ref{fig7}.C. Average grain area is accurately predicted by the network beyond pixel-level visual similarity, even when provided with fewer input frames.

\subsection{Spinodal Decomposition}

Spinodal decomposition describes spontaneous phase separation of a homogeneous mixture into compositionally distinct domains. In this work, the ground-truth evolution is generated by solving the CH equation~(\ref{CHS}), which couples interfacial energy with long-range diffusion and yields fourth-order nonlinear dynamics. Compared with grain growth, this diffusion-coupled evolution is more challenging to forecast, particularly over long horizons. The process typically exhibits a rapid early-time amplification of composition fluctuations followed by a slower coarsening stage, and we therefore assess prediction performance across both early and late evolution periods.

\subsubsection{Predictions of 90 future frames from 10 input frames (10$\rightarrow$90)}

\begin{figure}[h]
\centering
\includegraphics[width=0.9\linewidth]{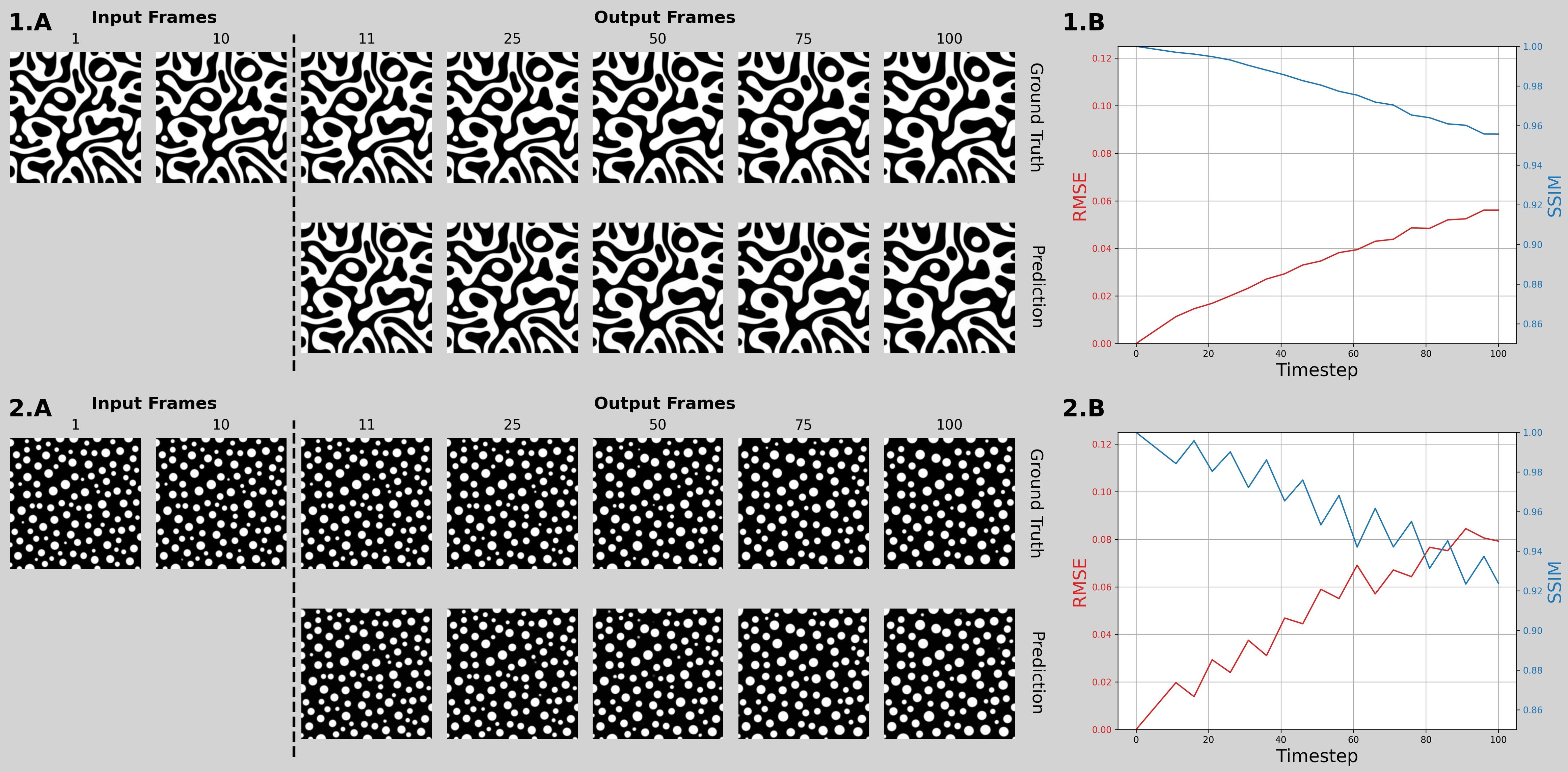}
\caption{\textbf{Spinodal decomposition prediction (10 input frames, 90 output frames) using the proposed fully convolutional spatiotemporal model:} Two test samples are shown  (indexed numerically). (A) Prediction sequence and the corresponding ground-truth sequence, shown at t = 11, 25, 50, 75, 100; (B) RMSE and SSIM computed between the predicted and ground-truth sequences at 5-frame intervals.}
\label{fig8}
\end{figure}

For the 10$\rightarrow$90 task, 600 test samples are evaluated. Two representative examples are shown in Fig.~\ref{fig8}. The input sequence is displayed at $t=1$ and $t=10$, while the predicted and ground-truth outputs are shown at $t=11, 25, 50, 75,$ and $100$ (Figs.~\ref{fig8}.1.A and \ref{fig8}.2.A). Both bicontinuous and droplet domain morphologies are accurately reproduced in the predicted sequences. The droplet structures, in particular, are spatially well-localized, with minor discrepancies observed along domain boundaries. RMSE and SSIM are computed at 5-frame intervals and presented in Figs.~\ref{fig8}.1.B and ~\ref{fig8}.2.B. For both samples, RMSE remains below 0.10 while SSIM exceeds 0.92, indicating strong predictive fidelity in capturing the temporal evolution of the ground-truth microstructures.

\begin{figure}[h]
\includegraphics[width=\linewidth]{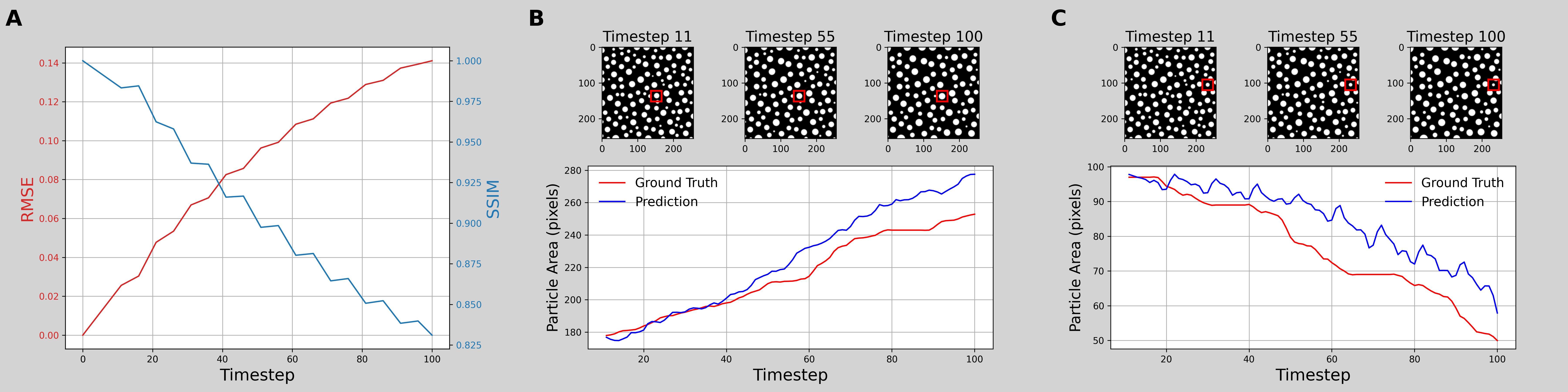}
\caption{\textbf{Accuracy metrics on spinodal decomposition prediction (10 input frames, 90 output frames):} (A) Dataset-averaged RMSE and SSIM evaluated every 5 frames over the prediction horizon. (B--C) The growth/shrinking of a select particle, boxed by a red square. The ground truth (red) and predicted (blue) are displayed together, measured at every frame.}
\label{fig9}
\end{figure}

Dataset-level performance and particle-size growth/shrink accuracy are summarized in Fig.~\ref{fig9}. Fig.~\ref{fig9}.A shows the mean RMSE and SSIM over the dataset, computed at 5-frame intervals, indicating consistently strong average predictive performance. Average RMSE just below 0.15 and average SSIM higher that 0.82 along the prediction horizon demonstrates a close relation between the prediction and ground-truth sequences. Fig.~\ref{fig9}.B compares the temporal evolution of a representative growing particle between the predicted and ground-truth sequences, while Fig.~\ref{fig9}.C presents the same analysis for a representative shrinking particle. Both figures are calculated from the particle sample in Fig.~\ref{fig8}.2.A. Ground-truth frames containing the tracked particle  -- displayed at $t = 11, 55,$ and $100$ -- with particle area quantified below at every timestep and Savitzky-Golay smoothing. In both figures, the particle is tracked via nearest-neighbor centroid association and highlighted with a red bounding box. The model accurately captures the rate of local morphological evolution, particularly during particle growth. Although fluctuations are observed in the shrinking-particle forecast, the predicted trajectory remains well-aligned with the ground truth, indicating robust modeling of particle-size coarsening dynamics.

\subsubsection{Predictions of 200 future frames from 10 input frames (10$\rightarrow$200)}

\begin{figure}[h!]
\centering
\includegraphics[width=0.9\linewidth]{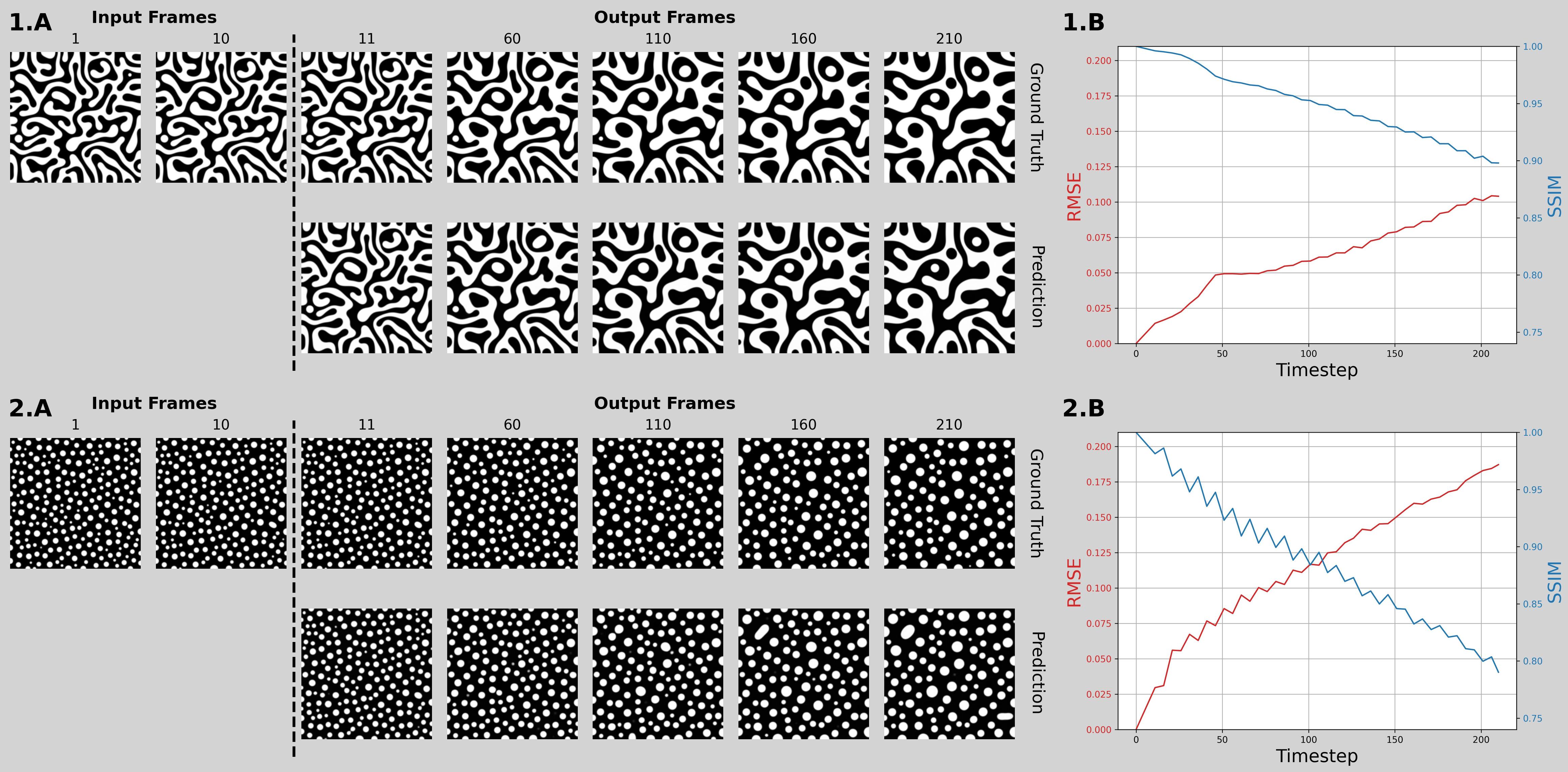}
\caption{\textbf{Spinodal decomposition prediction (10 input frames, 200 output frames) using the proposed fully convolutional spatiotemporal model:} Two test samples are shown  (indexed numerically). (A) Prediction sequence and the corresponding ground-truth sequence, shown at t = 11, 60, 110, 160, 210; (B) RMSE and SSIM computed between the predicted and ground-truth sequences at 5-frame intervals.}
\label{fig10}
\end{figure}

For the 10$\rightarrow$200 task, 50 test samples are evaluated using iterative roll-out to extend the prediction horizon beyond the nominal 90-frame output length. Two representative examples are shown in Fig.~\ref{fig10}. For compact visualization, the input sequence is displayed at $t=1$ and $t=10$, while the predicted and ground-truth outputs are shown at $t=11, 60, 110, 160,$ and $210$ (Figs.~\ref{fig10}.1.A and \ref{fig10}.2.A). Predictions closely match the ground-truth during the early stages, whereas deviations become more apparent at later forecast horizons. Along this extended prediction task, the network preserves physically consistent temporal relationships associated with microstructural coarsening, producing particularly accurate modeling in the central regions. Figs.~\ref{fig10}.1.B and \ref{fig10}.2.B report the RMSE and SSIM for each sample, computed at 5-frame intervals. The predictions achieve RMSE values below 0.20 and SSIM values exceeding 0.85, demonstrating strong overall performance on the long-range spinodal decomposition modeling task.

\begin{figure}[h!]
\includegraphics[width=\linewidth]{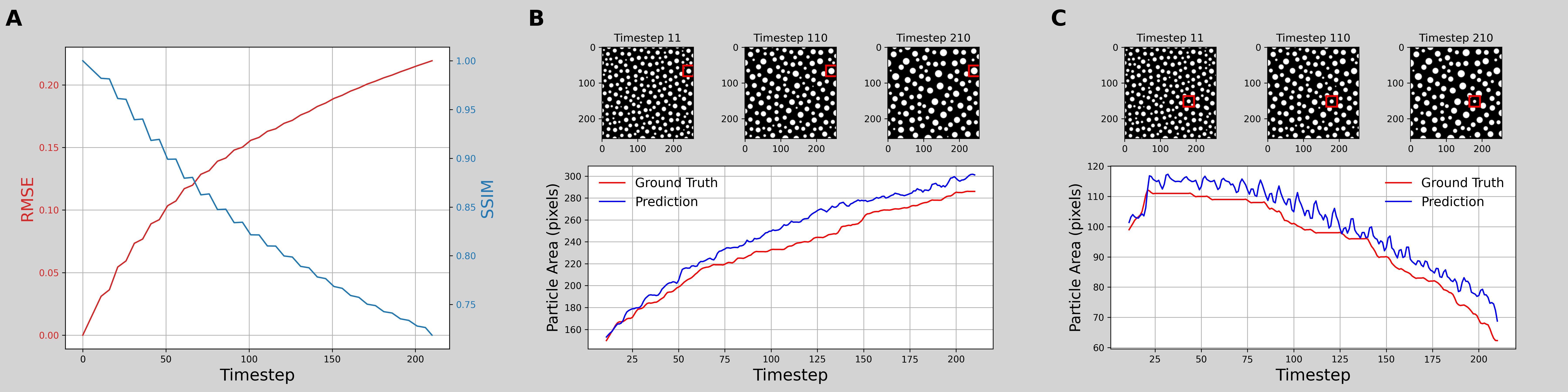}
\caption{\textbf{Accuracy metrics on spinodal decomposition prediction (10 input frames, 200 output frames):} (A) Dataset-averaged RMSE and SSIM evaluated every 5 frames over the prediction horizon. (B--C) The growth/shrinking of a select particle, boxed by a red square. The ground truth (red) and predicted (blue) are displayed together, measured at every frame.}
\label{fig11}
\end{figure}

Prediction performance is summarized along with particle size comparisons in Fig.~\ref{fig11}. Visual metric scores, RMSE and SSIM, are calculated at 5-frame intervals in Fig.~\ref{fig11}.A, displaying the model's strong predictive performance at early timesteps. Towards the later stages of prediction, average RMSE is below 0.25 and average SSIM at approximately 0.7, indicating maintained accuracy as the prediction horizon is extended through iterative roll-out. Figs.~\ref{fig11}.B--C present the morphological change of one particle growing and another particle shrinking in the particle sample shown in Fig.~\ref{fig10}.2.A. At the top of each figure, there are ground-truth frames displayed at $t = 11, 60, 110,$ and $210$, and the tracked particle is bounded by a red box. The particle is tracked by nearest-neighbor centroid association, and its evolution is quantified in the plot below at each timestep with Savitzky-Golay smoothing applied. The network accurately captures the rate at which the focused particles evolve over the prolonged extrapolation task. Both the growing and shrinking particles in the prediction closely model the dynamic temporal evolution of the ground truth, despite fluctuations.

\subsection{Computational Efficiency and Inference Speedup}
Compared with recurrent baselines, the proposed fully convolutional spatiotemporal model provides a substantial reduction in training and inference cost while maintaining predictive accuracy. As summarized in Table~\ref{tab:runtime}, for the grain growth task the proposed model requires only 30 seconds per inference, which corresponds to approximately 17.6$\times$ faster inference than ConvLSTM and 44.5$\times$ faster inference than PredRNN++ under the same evaluation setting. This speed advantage is consistent with the computational complexity: our model requires 9.327G Floating point operations per second (FLOPs), whereas ConvLSTM and PredRNN++ require 1.168T and 3.552T FLOPs, respectively. These results confirm that the fully convolutional design yields a significant practical improvement in deployment efficiency for microstructure evolution prediction, especially for long sequences and high-resolution evaluation.

\begin{table}[h!]
\centering
\begin{tabular}{|c|c|c|}
\hline
\textbf{Models} & \textbf{Inference Time (hh/mm/ss)} & \textbf{FLOPs} \\
\hline
The proposed model  & 00:00:30 & 9.327G \\
ConvLSTM  & 00:08:48 & 1.168T \\
PredRNN++ & 00:22:14 & 3.552T \\
\hline
\end{tabular}
\caption{\textbf{Inference efficiency comparison across deep spatiotemporal models on grain growth prediction}. Wall-clock inference time and estimated FLOPs for grain growth and spinodal decomposition prediction. The proposed fully convolutional model  achieves markedly lower inference time than recurrent baselines (ConvLSTM and PredRNN++), consistent with its substantially reduced FLOPs.}
\label{tab:runtime}
\end{table}

\section{Discussion and Conclusion}
\label{sec:discussion}
This work demonstrates that fully convolutional spatiotemporal learning models can accurately and efficiently predict the microstructure evolution generated by the phase-field method in two representative processes: grain growth and spinodal decomposition. The proposed framework avoids recurrent hidden-state propagation and instead learns evolution dynamics through lightweight convolutional operators with large effective receptive fields. This architectural choice yields two practical advantages observed consistently across our experiments: (i) stable long-horizon rollouts without catastrophic error growth, and (ii) substantially reduced inference cost compared with recurrent baselines.

For grain growth, the model reproduces the temporal evolution of grain-boundary networks and coarsening behavior with high fidelity. In the nominal 10$\rightarrow$90 setting, predictions remain visually consistent with ground truth over the entire horizon and preserve sharp grain interfaces, even when evaluated at $256\times256$ resolution after training only on $64\times64$ clips. Beyond pixel-wise similarity, physics-consistent behavior is supported by the GSD comparisons: the predicted GSDs closely track the ground truth at intermediate and late times, indicating that the model captures statistically meaningful coarsening trends rather than merely matching textures. The long-horizon 10$\rightarrow$190 experiments further highlight both the strength and the expected limitations of iterative roll-out. While RMSE increases and SSIM decreases as forecast time extends, the predicted sequences remain morphologically plausible and continue to preserve grain-boundary topology without introducing obvious artifacts. Importantly, the mean grain-area evolution and its linear trend remain close to the ground truth across the dataset, suggesting that the model maintains key coarsening signatures even when pixel-level errors accumulate. The limited-context 5$\rightarrow$95 setting demonstrates robustness when fewer observed frames are available than the network expects. Although zero-padding induces short-lived artifacts at early forecast times, the model rapidly transitions to stable predictions, and both RMSE/SSIM (evaluated after the padding transient) and GSD statistics indicate strong predictive performance. Together, these results suggest that the learned latent dynamics are not overly sensitive to the exact input length and can generalize under reduced temporal context.

For spinodal decomposition, the forecasting task is inherently more challenging due to the diffusion-coupled, fourth-order nonlinear dynamics described by the CH equation. Despite increased complexity, the proposed model captures both the qualitative evolution of phase-separated patterns and the temporal progression of coarsening, demonstrating that fully convolutional spatiotemporal learning can model diffusion-driven pattern formation beyond grain-boundary migration.

A particularly notable outcome is resolution generalization: models trained at low spatial resolution can be directly applied to substantially higher-resolution simulations without architectural modification or retraining. This capability follows naturally from the fully convolutional design and is empirically supported by the strong qualitative agreement and consistent morphology statistics at higher resolution. Such scalability is critical in practice because high-fidelity phase-field simulations at large spatial resolutions are computationally expensive, and surrogate predictors must remain reliable at resolutions relevant to downstream analysis and materials design.

In addition to predictive fidelity, the proposed approach provides a major improvement in computational efficiency. The runtime and FLOPs comparisons against ConvLSTM and PredRNN++ confirm that the fully convolutional formulation achieves order-of-magnitude inference speedups, consistent with substantially reduced FLOPs. This efficiency advantage is especially important for deployment scenarios that require repeated forecasting, parameter sweeps, uncertainty analysis, or integration into digital-twin pipelines where the predictor is invoked many times within an outer optimization or control loop.

Several limitations motivate future work. First, the current evaluation focuses on image-level and statistics-based fidelity (RMSE, SSIM, and GSD for grain growth). Incorporating additional physics-aware diagnostics, such as structure factor evolution for spinodal decomposition, interfacial length/curvature statistics, or free-energy decay consistency, would provide a more comprehensive assessment of physical plausibility. Second, while iterative roll-out enables long-horizon prediction, it inevitably accumulates error. Training strategies that explicitly optimize multi-step rollouts, incorporate scheduled sampling, or use consistency regularization across horizons may further improve long-term stability. Third, the present study trains separate models for grain growth and spinodal decomposition. A unified model conditioned on physics parameters or initial-state descriptors may enable broader generalization across multiple phase-field mechanisms and parameter regimes. Finally, integrating weak physics constraints (e.g., mass conservation for CH) or uncertainty quantification (e.g., probabilistic forecasts or ensemble prediction) could enhance reliability for downstream decision-making.

Overall, the results indicate that fully convolutional spatiotemporal models provide a compelling balance of accuracy, stability, and computational efficiency for microstructure evolution prediction, offering a scalable surrogate modeling foundation for accelerated phase-field simulation and digital-twin-enabled workflows in computational materials science.

\section*{Declaration of competing interest} 
The authors declare that they have no known competing financial interests or personal relationships that could have appeared to influence the work reported in this paper.

\section*{Declaration of generative AI and AI-assisted technologies in the manuscript preparation process} 
During the preparation of this work the author(s) used ChatGPT in order to improve language and readability. After using this tool/service, the author(s) reviewed and edited the content as needed and take(s) full responsibility for the content of the publication.

\section*{Acknowledgment}
X. Li's and S. M. Perera's work was partially funded by the Division of Mathematical Sciences, National Science Foundation with the award numbers 2410678 \& 2410676, respectively.

\bibliographystyle{elsarticle-num} 

\bibliography{reference}

\end{document}